\documentclass[runningheads]{llncs}
\usepackage{graphicx}
\usepackage{todonotes}
\usepackage{cite}

\begin{document}
%

\title{Automatic retrieval of corresponding US views in longitudinal examinations}

%



\authorrunning{F. Author et al.}                   

\author{Hamideh Kerdegari\inst{1}\thanks{This work was supported by the Wellcome Trust UK (110179/Z/15/Z, 203905/Z/16/Z, WT203148/Z/16/Z). H. Kerdegari, N. Phung, R. Razavi, A. P King and A. Gomez acknowledge financial support from the Department of Health via the National Institute for Health Research (NIHR) comprehensive Biomedical Research Centre award to Guy's and St Thomas' NHS Foundation Trust in partnership with King's College London and King's College Hospital NHS Foundation Trust.} \and
Tran Huy Nhat Phung\inst{1,3} \and
Van Hao Nguyen\inst{2} \and
Thi Phuong Thao Truong\inst{2} \and
Ngoc Minh Thu Le\inst{2} \and
Thanh Phuong Le\inst{2} \and
Thi Mai Thao Le\inst{2} \and
Luigi Pisani\inst{4} \and
Linda Denehy\inst{5} \and
Vital Consortium\inst{6} \and
Reza Razavi\inst{1} \and
Louise Thwaites\inst{3} \and
Sophie Yacoub\inst{3} \and
{Andrew P. King}\inst{1} \and
Alberto Gomez\inst{1}}


\authorrunning{H. Kerdegari et al.}
%
\institute{School of Biomedical Engineering \& Imaging Sciences, King’s College London, UK\and
Hospital for Tropical Diseases, Ho Chi Minh City, Vietnam \and
Oxford University Clinical Research Unit, Ho Chi Minh City, Vietnam \and
Mahidol Oxford Tropical Medicine Research Unit, Bangkok, Thailand \and
Melbourne School of Health Sciences, The University of Melbourne, Australia \and
Membership of the VITAL Consortium is provided in the Acknowledgments 
\email{hamideh.kerdegari@kcl.ac.uk}}
\maketitle      

\begin{abstract}
Skeletal muscle atrophy is a common occurrence in critically ill patients in the intensive care unit (ICU) who spend long periods in bed. Muscle mass must be recovered through physiotherapy before patient discharge and ultrasound imaging is frequently used to assess the recovery process by measuring the muscle size over time. However, these manual measurements are subject to large variability, particularly
since the scans are typically acquired on different days and potentially by different operators.
In this paper, we propose a self-supervised contrastive learning approach to automatically retrieve similar ultrasound muscle views at different scan times.
Three different models were compared using data from 67 patients acquired in the ICU. Results indicate that our contrastive model outperformed a
supervised baseline model in the task of view retrieval
with an AUC of 73.52\% and when combined with an automatic segmentation model achieved $5.7\%\pm0.24\%$ error in cross-sectional area. Furthermore, a user study survey confirmed the efficacy of our model for muscle view retrieval. 

\end{abstract}

\section{Introduction}

\begin{figure}[t]
\includegraphics[width=100mm, trim={0 50 0 0}, clip]{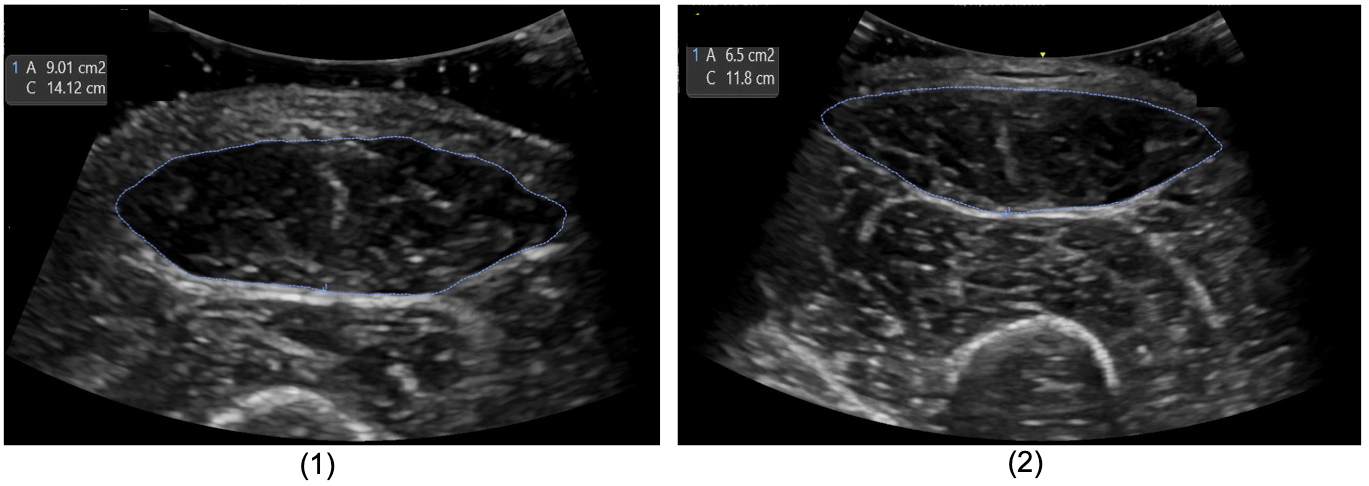}
\centering
\vspace{-1mm}
\caption{Example of the cross-section of the rectus femoris (RF) on one ICU patient showing muscle mass reduction from admission (9cm$^2$, left) to discharge (6cm$^2$, right).} \label{atrophy}
\end{figure}

Muscle wasting, also known as muscle atrophy (see Fig.~\ref{atrophy}), is a common complication in critically ill patients, especially in those who have been hospitalized in the intensive care unit (ICU) for a long period \cite{trung2019functional}. Factors contributing to muscle wasting in ICU patients include immobilization, malnutrition, inflammation, and the use of certain medications~\cite{puthucheary2013acute}. Muscle wasting can result in weakness, impaired mobility, and increased morbidity and mortality. Assessing the degree of muscle wasting in ICU patients is essential for monitoring their progress and tailoring their rehabilitation program to recover muscular mass through physiotherapy before patient discharge.
Traditional methods of assessing muscle wasting, such as physical examination, bioelectrical impedance analysis, and dual-energy X-ray absorptiometry, may be limited in ICUs due to the critical illness of patients~\cite{schefold2020muscular}. Instead, ultrasound (US) imaging has emerged as a reliable, non-invasive, portable tool for assessing muscle wasting in the ICU~\cite{mourtzakis2014bedside}.

The accuracy and reliability of US imaging in assessing muscle wasting in ICU patients have been demonstrated by Parry et al.~\cite{parry2015ultrasonography}. US imaging can provide accurate measurements of muscle size, thickness, and architecture, allowing clinicians to track changes over time. However, these measurements are typically performed manually, which is time-consuming, subject to large variability and depends on the expertise of the operator. Furthermore, operators might be different from day to day and/or start scanning from different positions in each scan which will cause further variability.

In recent years, self-supervised learning (SSL) has gained popularity for automated diagnosis in the field of medical imaging due to its ability to learn from unlabeled data  \cite{sowrirajan2021moco, hosseinzadeh2021systematic, azizi2021big, chen2021uscl}. Previous studies on SSL for medical imaging have focused on designing pretext tasks \cite{zhuang2019self, bai2019self, jiao2020self, hu2020self}. A class of SSL, contrastive learning (CL), aims to learn feature representations via a contrastive loss function to distinguish between negative and positive image samples. A relatively small number of works have applied CL to US imaging, for example to synchronize different cross-sectional views \cite{dezaki2021echo} and to perform view classification \cite{chartsias2021contrastive} in echocardiography (cardiac US).


In this paper, we focus on the underinvestigated application of view matching for longitudinal RF muscle US examinations to assess muscle wasting. Our method uses a CL approach (see Fig. 2) to learn a discriminative representation from muscle US data which facilitates the retrieval of similar muscle views from different scans.

The novel contributions of this paper are: 1) the first investigation of the problem of muscle US view matching for longitudinal image analysis, and 2) our approach is able to automatically retrieve similar muscle views between different scans, as shown by quantitative validation and qualitatively through a clinical survey.
\begin{figure}[t]
\includegraphics[width=\textwidth, trim={0 0 0 13}, clip]{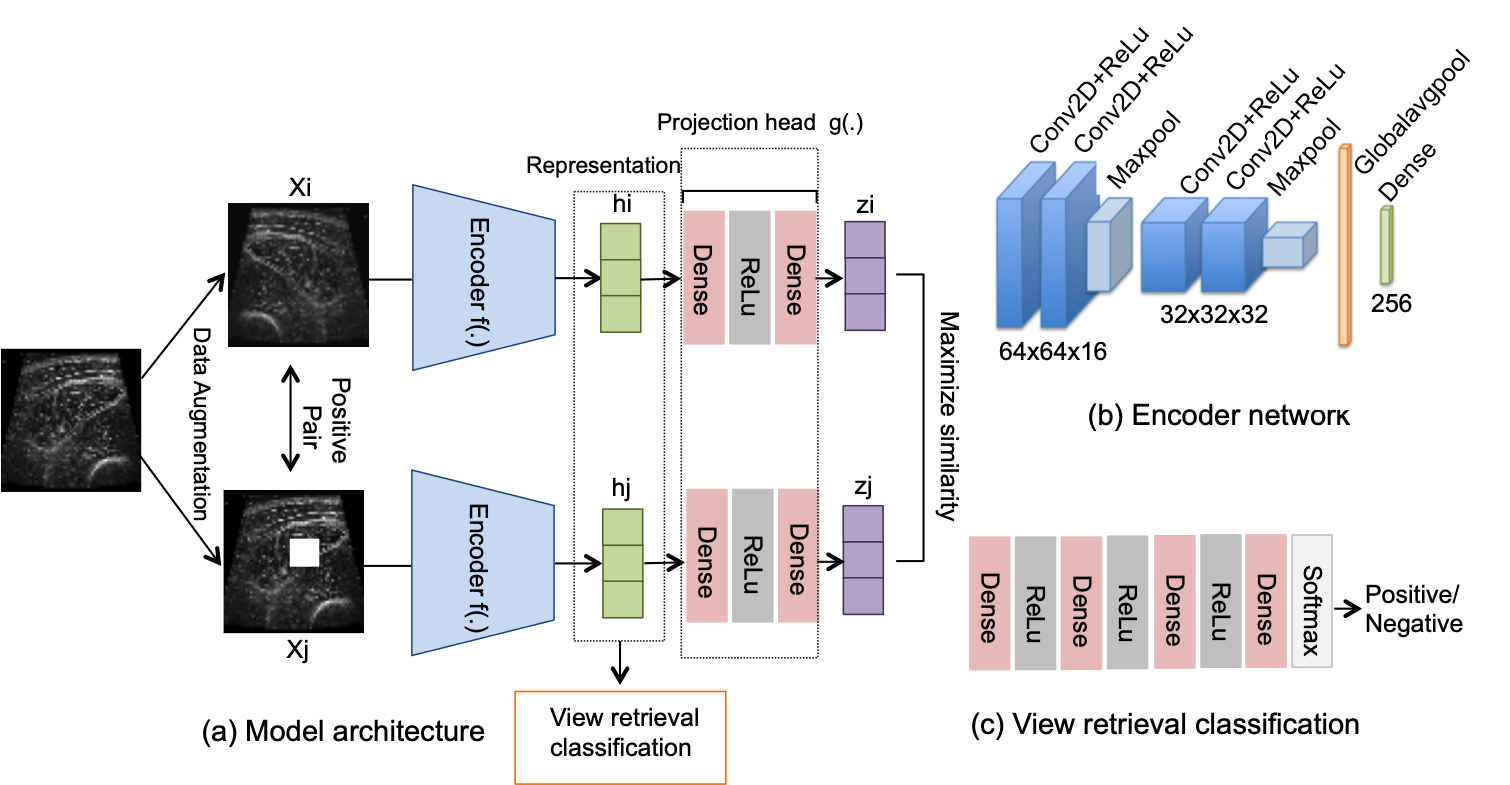}
\caption{Proposed architecture for US view retrieval. (a): Overview, a shared encoder and a projection head (two dense layers, each 512 nodes). (b): Encoder subnetwork. (c): The classification subnetwork has four dense layers of 2024, 1024, 512 and 2 features.} \label{fig3}
\end{figure}

\section{Method}

\subsection{Problem Formulation}
Muscle wasting assessment requires matching of corresponding cross-sectional US views of the RF over subsequent (days to weeks apart) examinations. The first acquisition is carried out following a protocol to place the transducer half way through the thigh and perpendicular to the skin, but small variations in translation and angulation  away from this standard view are common. This scan produces the reference view at time $T_1$ ($RT_1$). The problem is as follows: given $RT_1$, the task is to retrieve the corresponding view ($VT_2$) at a later time ($T_2$) from a sequence of US images captured by the operator using the transducer at approximately the same location and angle as for $T_1$. The main challenges of this problem include: (1) the transducer pose and angle might be different, (2) machine settings might be slightly different, and (3) parts of the anatomy (specifically the RF) might change in shape and size over time. As a result, our aim is to develop a model that can select the most similar view acquired during $T_2$ to the reference view $RT_1$ acquired at $T_1$.


\subsection{Contrastive Learning Framework for Muscle View Matching}


Inspired by the SimCLR algorithm \cite{chen2020simple}, our model learns representations by maximizing the similarity between two different augmented views of the same muscle US image via a contrastive loss in the latent space. We randomly sample a minibatch of $N$ images from the  video sequences over three times $T_1$, $T_2$ and $T_3$, and define the contrastive learning on positive pairs $(Xi, Xj)$ of augmented images derived from the minibatch, resulting in $2N$ samples. Rather than explicitly sampling negative examples, given a positive pair, we consider the other $2(N-1)$ augmented image pairs within a minibatch as negative.

The contrastive loss function for a positive pair $(Xi, Xj)$ is defined as:

\begin{equation}
L^{i}_{C}= -log\frac{exp(sim(z_{i},z_{j})/\tau )_{}}{\sum_{k=1}^{2n} 1_{[k\neq i]}exp(sim(z_{i},z_{k})/\tau )},
\end{equation}

where $1\in(0,1)$, $\tau$ is a temperature parameter and $sim(\cdot)$ denotes the pairwise cosine similarity. $z$ is a representation vector, calculated by $z = g(f(X))$, where f(·) indicates a shared encoder and g(·) is a projection head. $L^{i}_{C}$ is computed across all positive pairs in a mini-batch. Then $f(\cdot)$ and $g(\cdot)$ are trained to maximize similarity using this contrastive loss.


\subsection{The Model Architecture}

The model architecture is shown in Fig. \ref{fig3}a. First, we train the contrastive model to identify the similarity between two images, which are a pair of image augmentations created by horizontal flipping and random cropping (size 10$\times$10) applied on a US image (i.e., they represent different versions of the same image). Each image of this pair $(Xi, Xj)$ is fed into an encoder to extract representation vectors $(hi, hj)$ from them. The encoder architecture (Fig. \ref{fig3}b) has four conv layers (kernel $3\times3$) with ReLU and two max-poolings. A projection head (a multilayer perceptron with two dense layers of 512 nodes) follows  mapping these representations to the space where the contrastive loss is applied.

Second, we use the trained encoder $f(\cdot)$ for the training of our main task (i.e. the downstream task), which is the classification of positive and negative matches (corresponding and non-corresponding views) of our test set. For that, we feed a reference image $X_{ref}$, and a candidate frame $X_{j}$ to the encoder to obtain the representations $hi$, $hj$ and feed these in turn to a classification network (shown in Fig. \ref{fig3}c) that contains four dense layers with ReLU activation and a softmax layer.


\section{Materials}
The muscle US exams were performed using GE Venue Go and GE Vivid IQ machines, both with linear probes (4.2-13.0 MHz), by five different. During examination, patients were in supine position with the legs in a neutral rotation with relaxed muscle and passive extension. Measurements were taken at the point three fifths of the way between the anterior superior iliac spine and the patella upper pole.
The transducer was placed perpendicular to the skin and to the longitudinal axis of the thigh to get the cross-sectional area of the RF. An excess of US gel was used and pressure on the skin was kept minimal to maximise image quality. US measurements were taken at ICU admission ($T_1$), 2-7 days after admission ($T_2$) and at ICU discharge ($T_3$).
For this study, 67 Central Nervous System (CNS) and Tetanus patients were recruited and their data were acquired between June 2020 and Feb 2022. Each patient had an average of six muscle ultrasound examinations, three scans for each leg, totalling 402 examinations. The video resolution was 1080 × 1920 with a frame rate of 30fps. This study was performed in line with the principles of the Declaration of Helsinki. Approval was granted by the Ethics Committee of the Hospital for Tropical Diseases, Ho Chi Minh City and Oxford Tropical Research Ethics Committee. 

\begin{figure}[t]
\includegraphics[width=80mm]{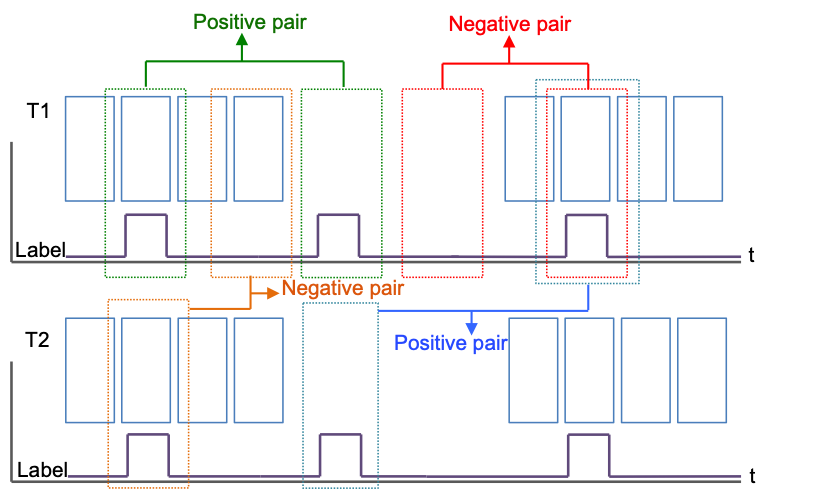}
\centering
\vspace{-1mm}
\caption{An example of positive and negative pair labeling for US videos acquired at $T_1$ and $T_2$. Positive pairs are either the three views acquired consecutively at the $T_i$, or a view labeled at $T_1$ and the corresponding view on the same leg at $T_2$ or $T_3$.} 
\label{label}
\end{figure}

The contrastive learning network was trained without any annotations. However, for the view matching classification task, our test data were annotated automatically 
as positive and negative pairs based upon manual frame selection by a team of five doctors comprising three radiologists and two ultrasound specialists with expertise in muscle ultrasound. Specifically, each frame in an examination was manually labelled as containing a similar view to the reference $RT_1$ or not. Based upon these labellings, as shown in Fig. \ref{label}, the positive pairs are combinations of similar views within each examination $(T_1/T_2/T_3)$ and between examinations. The rest are considered negative pairs.

\section{Experiments and Results}

\subsection{Implementation Details}
Our model was implemented using Tensorflow 2.7. During training, input videos underwent experimentation with clip sizes of $256 \times 256$, $128 \times 128$, and $64 \times 64$. Eventually, they were resized to $64 \times 64$ clips, which yielded the best performance. All the hyperparameters were chosen using the validation set. For the CL training, the standard Adam optimizer was used with learning rate =0.00001, kernel size = $3 \times 3$, batch size = 128, batch normalization, dropout with p = 0.2 and L2 regularization of the model parameters with a weight = 0.00001. The CL model was trained on 80\% of the muscle US data for 500 epochs. For the view retrieval model, the standard Adam optimizer with learning rate = 0.0001, batch size = 42 and dropout of p = 0.2 was used. The classifier was trained on the remaining 20\% of the data (of which 80\% were used for training, 10\% for validation and 10\% for testing) and the
network converged after 60 epochs. For the supervised baseline model, the standard Adam optimizer was used with learning rate =0.00001, kernel size = $3 \times 3$, batch size = 40, and batch normalization. Here, we used the same data splitting as our view retrieval classifier. The code we used to train and evaluate our models is available at \url{https://github.com/hamidehkerdegari/Muscle-view-retrieval}.


\subsection{Results}

\subsubsection{Quantitative Results}
We carried out two quantitative experiments. First, we evaluated the performance of the view classifier. Second, we evaluated the quality of the resulting cross-sectional areas segmented using a U-Net \cite{ronneberger2015u}.

The classifier performance was carried out by measuring, for the view retrieval task, the following metrics: Area Under the Curve (AUC), precision, recall, and F1-score. Because there is no existing state of the art for this task, we created two  baseline models to compare our proposed model to: first, a naive image-space comparison using normalized cross-correlation (NCC) \cite{bourke1996cross}, and second, a supervised classifier. The supervised classifier has the same architecture as our CL model, but with the outputs of the two networks being concatenated after the representation $h$ followed by a dense layer with two nodes and a softmax activation function to produce the probabilities of being a positive or negative pair. 
Table~\ref{tab2} shows the classification results on our dataset.


\begin{table}
\caption{AUC, precision, recall and F1 score results on the muscle video dataset.}\label{tab2}
\label{tab:Accuracy}
\centering
\begin{tabular}{l| c c c c}
\hline
\hline
\textbf{Model}                                                             & \textbf {AUC   }& \textbf{Precision   } & \textbf{Recall   } & \textbf{F1}\\ \hline
Normalized cross-correlation                                                                       &      68.35 \%   & 58.65 \% &     63.12 \%&   60.8 \%  \\ \hline
Supervised baseline model                                                                   &    69.87 \% &    65.81 \%& 60.57 \%& 63.08 \%  \\ \hline
\textbf{Proposed model}       &    \textbf{73.52 \%}   &    \textbf{67.2 \%} &  \textbf{68.31 \%} &  \textbf{67.74 \%}      \\ \hline
\hline
\end{tabular}
\end{table}

\begin{figure}[h]
\includegraphics[width=110mm]{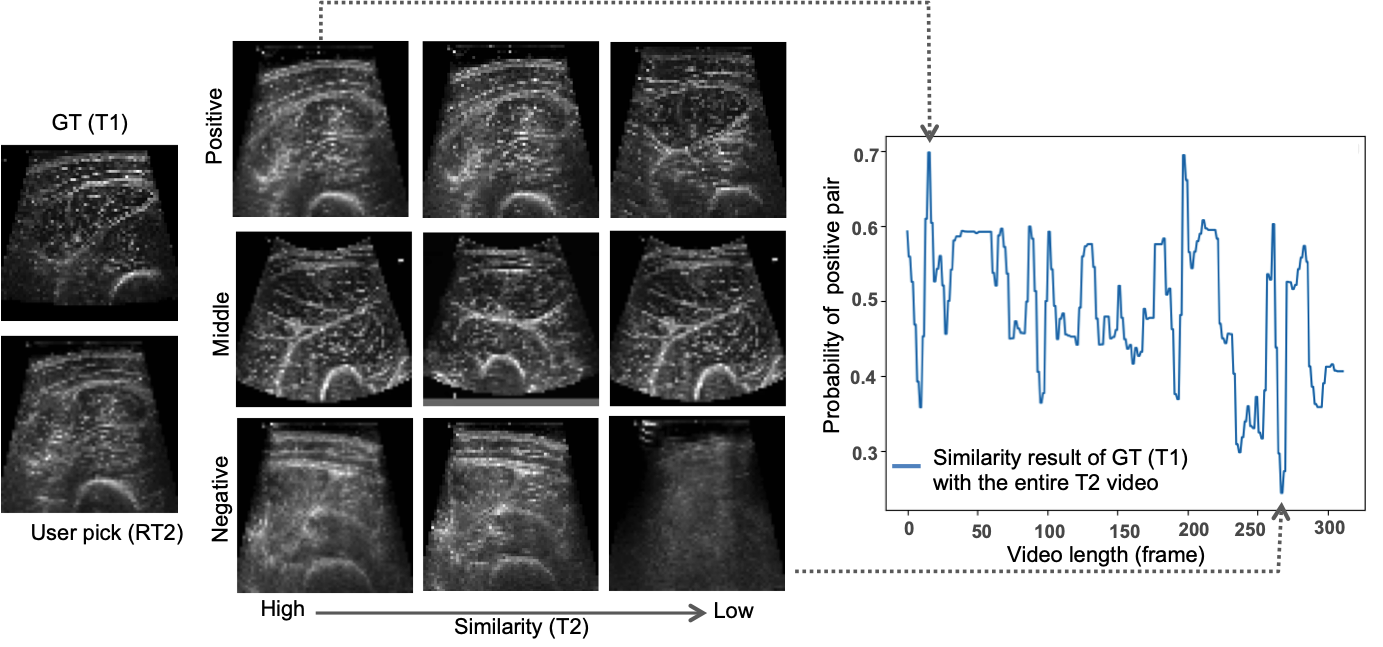}
\vspace{-2mm}
\centering
\caption{Results showing three sample positive, medium and negative predicted pairs by our model when ground truth (GT) from $T_1$ is compared with the $T_2$ video.} \label{fig2}
\end{figure}

As shown in Table~\ref{tab2}, our proposed method achieved superior performance in terms of AUC, precision, recall, and F1-score compared to all other models. The NCC method demonstrated the lowest performance, as it lacked the capability to accurately capture dynamic changes and deformations in US images which can result in significant structural differences. A representative example of a model-retrieved view for one case is presented in Fig. \ref{fig2}. It shows positive, 
negative, and middle (i.e., images with a probability value between the highest and lowest values predicted by our model) pairs of images generated by our model from a patient's left leg. As reference, on the left we show the user pick ($RT_2$). 

\begin{figure}
\includegraphics[width=120mm, trim={0 0 0 0}, clip]{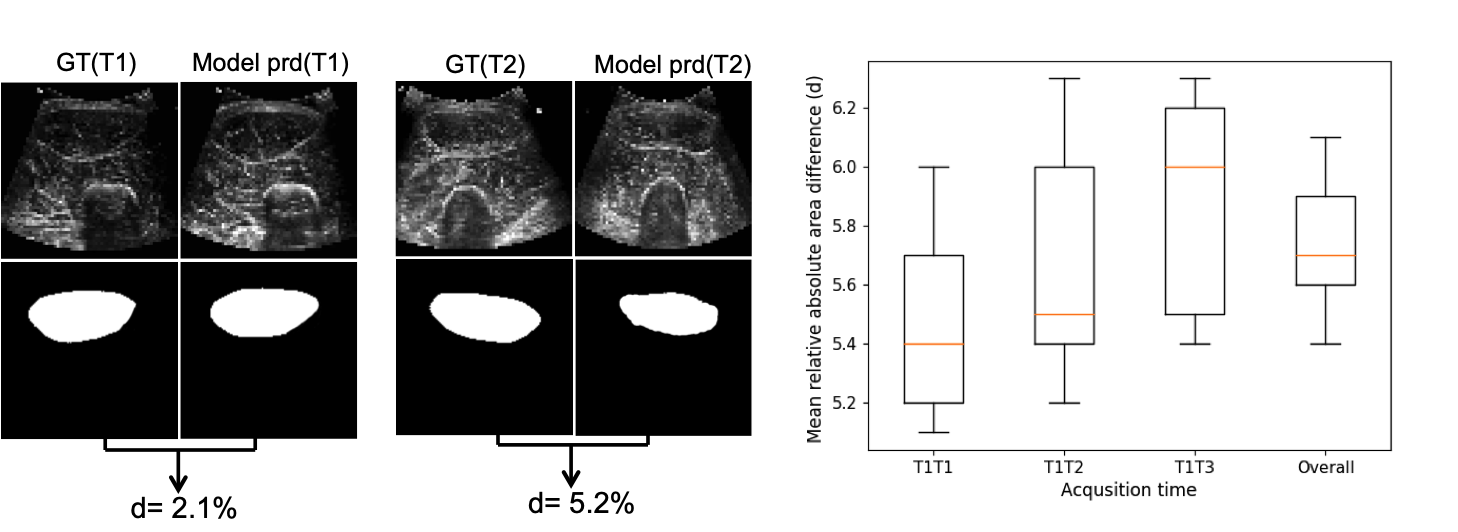}
\centering
\caption{Left: cross-sectional area error for $T_1$ and $T_2$ examinations (acquisition times). Right: mean relative absolute area difference (d) for $T_1T_1$, $T_1T_2$, $T_1T_3$ (reference frame from T1 and corresponding predicted frames from T1, T2 and T3 respectively) and overall acquisition time.} \label{fig4}
\end{figure}

To assess the quality of the resulting cross-sections, we calculated the mean relative absolute area difference ($d$) between the ground truth ($a_{GT}$) frame and that of the model predicted frame ($a_{pred}$) for each examination as follows:
\begin{equation}
d =\frac{|a_{GT}-a_{pred}|}{a_{GT}}\
\end{equation}
We applied a trained U-Net model (already trained with 1000 different US muscle images and manual segmentations). Results showed an overall cross-sectional mean relative absolute area error of $5.7\%\pm0.24\%$ on the test set (Full details provided in Fig. \ref{fig4}, right). To put this number into context, Fig. \ref{fig4}, left visualizes two cases where the relative error is 2.1\% and 5.2\%.

\subsubsection{Qualitative Results}

\begin{figure}
\centering
\includegraphics[width=120mm, trim={0 10 0 0}, clip]{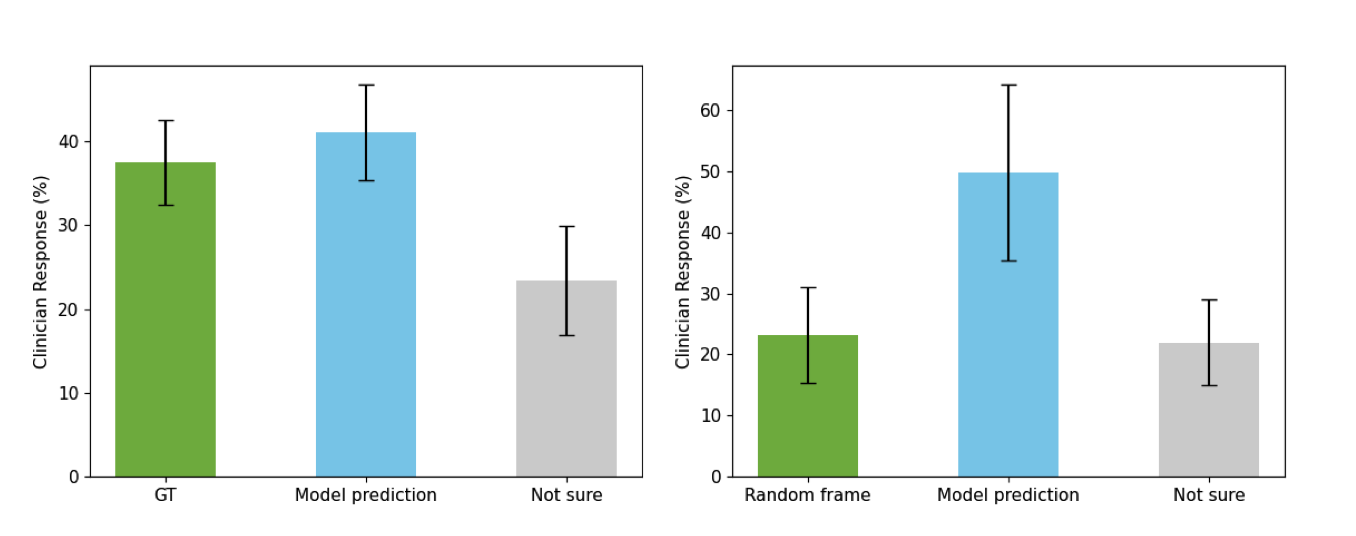}
\vspace{-2mm}
\caption{User study survey results. Left: when $T_1$GT, $T_2$GT and model prediction (from $T_2$) are shown to the users. Right: when $T_1$GT, $T_2$-random frame and model prediction (from $T_2$) are shown to the users.} \label{fig5}
\end{figure}

We conducted a user study survey to qualitatively assess our model's performance. The survey was conducted blindly and independently by four clinicians and consisted of thirty questions.
In each, clinicians were shown two different series of three views of the RF: (1) $RT_1$, GT match from $T_2$ and model prediction from $T_2$, and (2) $RT_1$, a random frame from $T_2$ and model prediction from $T_2$. They were asked to indicate which (second or third) was the best match with the first image.  
The first question aimed to determine if the model's performance was on par with clinicians, while the second aimed to determine if the model's selection of images was superior to a randomly picked frame. As shown in Fig. \ref{fig5}, left, clinicians chose the model prediction more often than the GT; however, this difference was not significant (paired Student's $t$-test, $p=0.44$, significance$=0.05$). Therefore, our model can retrieve the view as well as clinicians, and significantly better (Fig. \ref{fig5}, right) than randomly chosen frames (paired Student's $t$-test, $p=0.02$, significance$=0.05$).

\section{Discussion and Conclusion}
This paper has presented a self-supervised CL approach for automatic muscle US view retrieval in ICU patients. We trained a classifier to find positive and negative matches. We also computed the cross-sectional area error between the ground truth frame and the model prediction in each acquisition time to evaluate model performance. The performance of our model was evaluated on our muscle US video dataset and showed AUC of 73.52\% and $5.7\%\pm0.24\%$ error in cross-sectional view. Results showed that our model outperformed the supervised baseline approach. This is the first work proposed to identify corresponding ultrasound views over time, addressing an unmet clinical need.


\section{Acknowledgments}
\label{sec:acknowledgments}
The VITAL Consortium: \textbf{OUCRU}: Dang Phuong Thao, Dang Trung Kien, Doan Bui Xuan Thy, Dong Huu Khanh Trinh, Du Hong Duc, Ronald Geskus, Ho Bich Hai, Ho Quang Chanh, Ho Van Hien, Huynh Trung Trieu, Evelyne Kestelyn, Lam Minh Yen, Le Dinh Van Khoa, Le Thanh Phuong, Le Thuy Thuy Khanh, Luu Hoai Bao Tran, Luu Phuoc An, Nguyen Lam Vuong, Ngan Nguyen Lyle, Nguyen Quang Huy, Nguyen Than Ha Quyen, Nguyen Thanh Ngoc, Nguyen Thi Giang, Nguyen Thi Diem Trinh, Nguyen Thi Kim Anh, Nguyen Thi Le Thanh, Nguyen Thi Phuong Dung, Nguyen Thi Phuong Thao, Ninh Thi Thanh Van, Pham Tieu Kieu, Phan Nguyen Quoc Khanh, Phung Khanh Lam, Phung Tran Huy Nhat, Guy Thwaites, Louise Thwaites, Tran Minh Duc, Trinh Manh Hung, Hugo Turner, Jennifer Ilo Van Nuil, Vo Tan Hoang, Vu Ngo Thanh Huyen, Sophie Yacoub. \textbf{Hospital for Tropical Diseases, Ho Chi Minh City}: Cao Thi Tam, Ha Thi Hai Duong, Ho Dang Trung Nghia, Le Buu Chau, Le Mau Toan, Nguyen Hoan Phu, Nguyen Quoc Viet, Nguyen Thanh Dung, Nguyen Thanh Nguyen, Nguyen Thanh Phong, Nguyen Thi Cam Huong, Nguyen Van Hao, Nguyen Van Thanh Duoc,  Pham Kieu Nguyet Oanh, Phan Thi Hong Van, Phan Vinh Tho, Truong Thi Phuong Thao. \textbf{University of Oxford}: Natasha Ali, James Anibal, David Clifton, Mike English, Ping Lu, Jacob McKnight, Chris Paton, Tingting Zhu \textbf{Imperial College London}: Pantelis Georgiou, Bernard Hernandez Perez, Kerri Hill-Cawthorne, Alison Holmes, Stefan Karolcik, Damien Ming, Nicolas Moser, Jesus Rodriguez Manzano. \textbf{King’s College London}: Liane Canas, Alberto Gomez, Hamideh Kerdegari, Andrew King, Marc Modat, Reza Razavi. \textbf{University of Ulm}: Walter Karlen. \textbf{Melbourne University}: Linda Denehy, Thomas Rollinson. \textbf{Mahidol Oxford Tropical Medicine Research Unit (MORU)}: Luigi Pisani, Marcus Schultz
%
%
%
\bibliographystyle{splncs04}
\bibliography{ref}


\end{document}